\documentclass[letterpaper, 10 pt, journal, twoside]{IEEEtran}
\usepackage{amsmath,amsfonts}
\usepackage{algorithmic}
\usepackage{algorithm}
\usepackage{color}
\usepackage{array}
\usepackage{threeparttable}
\usepackage[caption=false,font=normalsize,labelfont=sf,textfont=sf]{subfig}
\usepackage{textcomp}
\usepackage{amssymb}
\usepackage{stfloats}
\usepackage{url}
\usepackage{verbatim}
\usepackage{graphicx}
\usepackage{cite}
\usepackage{multirow}
\usepackage{threeparttable}

\begin{document}

\title{Autonomous Robotic Bone Micro-Milling System with Automatic Calibration and 3D Surface Fitting
}

\author{Enduo Zhao, Xiaofeng Lin, Yifan Wang, and Kanako Harada, \IEEEmembership{Member,~IEEE}
\thanks{Manuscript received: June 30, 2025; revised: November 11, 2025; accepted: December 5, 2025. This letter was recommended for publication by Editor J. Burgner-Kahrs upon evaluation of the Associate Editor and Reviewers’ comments. This work was supported by JST Moonshot R\&D JPMJMS2033. \emph{(Corresponding author: Xiaofeng Lin.)}}
\thanks{Enduo Zhao was with the Department of Mechanical Engineering, Graduate School of Engineering, The University of Tokyo, Tokyo, Japan, when this work was conducted. He is currently with the School of Biomedical Engineering, Tsinghua University, Beijing, China. (e-mail: endowzhao@mail.tsinghua.edu.cn.)}%
\thanks{Xiaofeng Lin and Kanako Harada are with the Center for Disease Biology and Integrative Medicine, Graduate School of Medicine, The University of Tokyo, Tokyo, Japan. (email: \{lin-xiaofeng, kanakoharada\}@g.ecc.u-tokyo.ac.jp.)}%
\thanks{Yifan Wang is with the Department of Mechanical Engineering, Graduate School of Engineering, The University of Tokyo, Tokyo, Japan. (email: wang-yifan971125@g.ecc.u-tokyo.ac.jp.)}%
\thanks{Digital Object Identifer (DOI): see top of this page.}
}

\markboth{IEEE Robotics and Automation Letters. Preprint Version. Accepted DECEMBER, 2025}
{Enduo Zhao \MakeLowercase{\textit{et al.}}: Autonomous Robotic Bone Micro-Milling System} 

\maketitle

\begin{abstract}

Automating bone micro-milling using a robotic system presents challenges due to the uncertainties in both the external and internal features of bone tissue. For example, during mouse cranial window creation, a circular path with a radius of 2 to 4 mm needs to be milled on the mouse skull using a microdrill. The uneven surface and non-uniform thickness of the mouse skull make it difficult to fully automate this process, requiring the system to possess advanced perceptual and adaptive capabilities. In this study, we address this challenge by integrating a Microscopic Stereo Camera System (MSCS) into the robotic bone micro-milling system and proposing a novel online pre-measurement pipeline for the target surface. Starting from uncalibrated cameras, the pipeline enables automatic calibration and 3D surface fitting through a convolutional neural network (CNN)-based keypoint detection. Combined with the existing feedback-based system, we develop the world's first autonomous robotic bone micro-milling system capable of rapidly, in real-time perceiving and adapting to surface unevenness and non-uniform thickness, thereby enabling an end-to-end autonomous cranial window creation workflow without human assistance. Validation experiments on euthanized mice demonstrate that the improved system achieves a success rate of 85.7\,\% and an average milling time of 2.1 minutes, showing not only significant performance improvements over the previous system but also exceptional accuracy, speed, and stability compared to human operators.

\end{abstract}

\begin{IEEEkeywords}
Robotics and Automation in Life Sciences, Computer Vision for Medical Robotics, Visual Servoing, Computer Vision for Automation.
\end{IEEEkeywords}

\section{Introduction}\label{sec:introduction}

\IEEEPARstart{B}{one} micro-milling is an essential biological operation that involves using specialized milling tools to precisely remove or shape bone tissue at a microscale level \cite{liu2019fracture}. This demands a high level of skill from the human operator due to the tiny scale and the need for high accuracy and safety. Automating this process with a robotic system could, in principle, make it more accurate and reproducible than human operation. However, this presents challenges for the perceptual and adaptive capabilities of the robotic system. The system must first accurately perceive the uncertainties in the characteristics of bone tissue, including external features such as surface flatness, as well as internal features like thickness and heterogeneity \cite{shang2023cutting}, and generate milling trajectories that adapt to these uncertainties.

\begin{figure}[!t]
\centering
\includegraphics[width=2.7in]{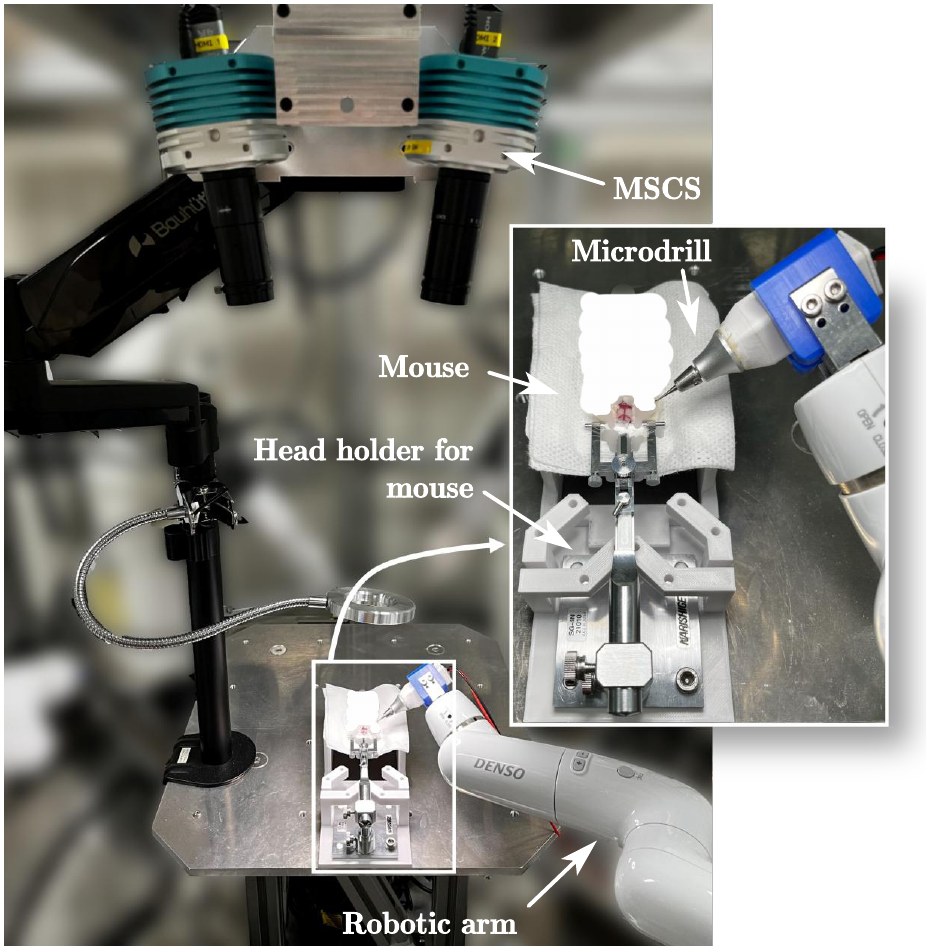}
\caption{The hardware setup of the autonomous robotic bone micro-milling system, including a robotic arm holding a microdrill, a head holder for the mouse to be milled, and a Microscopic Stereo Camera System (MSCS).}
\label{fig:harware}
\end{figure}

For instance, in scientific experiments involving the transplantation of human organoids into mice and monitoring their growth \cite{takebe2019}, an important preparatory step is creating a 2–4 mm circular cranial window on the mouse skull, whose thickness varies from 0.27 to 0.51 mm depending on strain, sex, and age. To avoid damaging the fragile underlying membrane, skilled operators use a handheld microdrill under a high-magnification microscope to mill along a circular trajectory, ensuring uniform remaining thickness, and then remove the central bone flap with tweezers. This procedure demands high operator skill, making automation essential for improving repeatability and robustness.

Recent studies have made significant progress in semi- or fully-automated craniotomy in medium-sized mammals. Xu et al. \cite{xu2021preliminary} developed a robotic-assisted system integrating human–robot collaboration and safe motion constraints, validated on Beagle models. Bian et al. \cite{bian2024automatic} proposed a feed-force-monitoring algorithm for autonomous cranium milling, tested on in vitro dog and goat specimens. Zhichao et al. \cite{zhichao2025precise} introduced a computed tomography (CT)-based path planning method using the Virtual-Center approach, validated on living dogs. Compared with these animals, the mouse skull is significantly thinner and softer, with higher sensitivity to surface unevenness and non-uniform thickness. These characteristics impose stricter requirements on robotic precision, stability and safety, and potential skull deformation during milling introduces dynamic challenges for fully autonomous operation.

To address these challenges in mice, several approaches have attempted to employ offline measurement, which involves acquiring the skull anatomical information prior to milling through preoperative imaging or manual surface scanning. Pak et al. \cite{pak2015} used contact scanning, Ghanbari et al. \cite{ghanbari2019} applied micro-CT, Navabi et al. \cite{navabi25} combined optical coherence tomography (OCT) with machine learning, and Jeong et al. \cite{jeong2013all} utilized ultrashort pulsed Ti-sapphire lasers for skull metrology. However, these offline measurement–based methods provide only static anatomical information of the skull, lack real-time feedback to adapt to dynamic changes, and require time-consuming measurements for each specimen, inevitably prolonging the overall operation time.

Other studies have attempted to achieve autonomous cranial window creation in mice using intraoperative feedback for real-time adaptation. Jia et al. \cite{jia2024helical} employed Bayesian transfer learning from manual eggshell milling to eggs of different sizes using images and force as feedback. In our previous work \cite{zhao2023}, we used CNN-based image feedback, achieving an 80\% success rate and a milling time of 16.8 minutes in eggshell milling. Due to ethical constraints, these systems were validated only on eggshells, whose thickness ($\approx 0.4  \ \mathrm{mm}$) and underlying delicate membrane are similar to those of the mouse skull \cite{andreoli2018egg}. However, in these feedback-based methods, the lack of pre-measured skull anatomical information prevents the initial trajectory from being pre-fitted to the target surface, requiring manual adjustment via teleoperation, which increases idle time, prolongs the overall milling duration, and raises the risk of failure. Moreover, the limited duration of inhalation anesthesia in mice (20–30 minutes \cite{oh2024mouse}), the more complex structural characteristics of the mouse skull compared to eggshells, and the fragile underlying brain tissue make it difficult to directly apply these methods to live mice, where higher milling speed and success rate are required.

\begin{table*}[]
\centering
\begin{threeparttable}
\caption{Capabilities of this study compared to existing works.}\label{table:1}
\begin{tabular}{|c|c||c|c|c|c|c|c|c|c|c|}
\hline
                         & This study & \cite{xu2021preliminary} & \cite{bian2024automatic} & \cite{zhichao2025precise}  & \cite{pak2015} & \cite{ghanbari2019} & \cite{navabi25} & \cite{jeong2013all} & \cite{jia2024helical} & \cite{zhao2023} \\ \hline \hline
Pre-measurement\tnote{1}      & \checkmark          & \checkmark       & \checkmark       & \checkmark       & \checkmark       & \checkmark       & \checkmark        & \checkmark        & ×        & ×        \\ \hline
Real-time feedback       & \checkmark          & \checkmark       & \checkmark       & ×       & ×       & ×       & ×        & ×        & \checkmark        & \checkmark        \\ \hline
Short overall duration   & \checkmark          & ×       & \checkmark       & ×       & ×       & ×       & \checkmark        & \checkmark        & ×        & ×        \\ \hline
Without human assistance\tnote{2} & \checkmark          & ×       & ×       & ×       & ×       & \checkmark       & ×        & \checkmark        & ×        & ×        \\ \hline
Validation on mouse      & \checkmark          & ×       & ×       & ×       & \checkmark       & \checkmark       & \checkmark        & \checkmark        & ×        & ×        \\ \hline
\end{tabular}

\begin{tablenotes}
\footnotesize
\item[1] Pre-measurement refers to obtaining skull anatomical information via offline measurement in \cite{xu2021preliminary,bian2024automatic,zhichao2025precise,pak2015,ghanbari2019,navabi25,jeong2013all}, or via online measurement in this study.
\item[2] Human assistance including scanning process in \cite{xu2021preliminary,bian2024automatic,zhichao2025precise,pak2015,navabi25}, or initial trajectory adjustment in \cite{jia2024helical,zhao2023}.
\end{tablenotes}
\end{threeparttable}
\end{table*}

\begin{figure}[!t]
\centering
\includegraphics[width=3.2in]{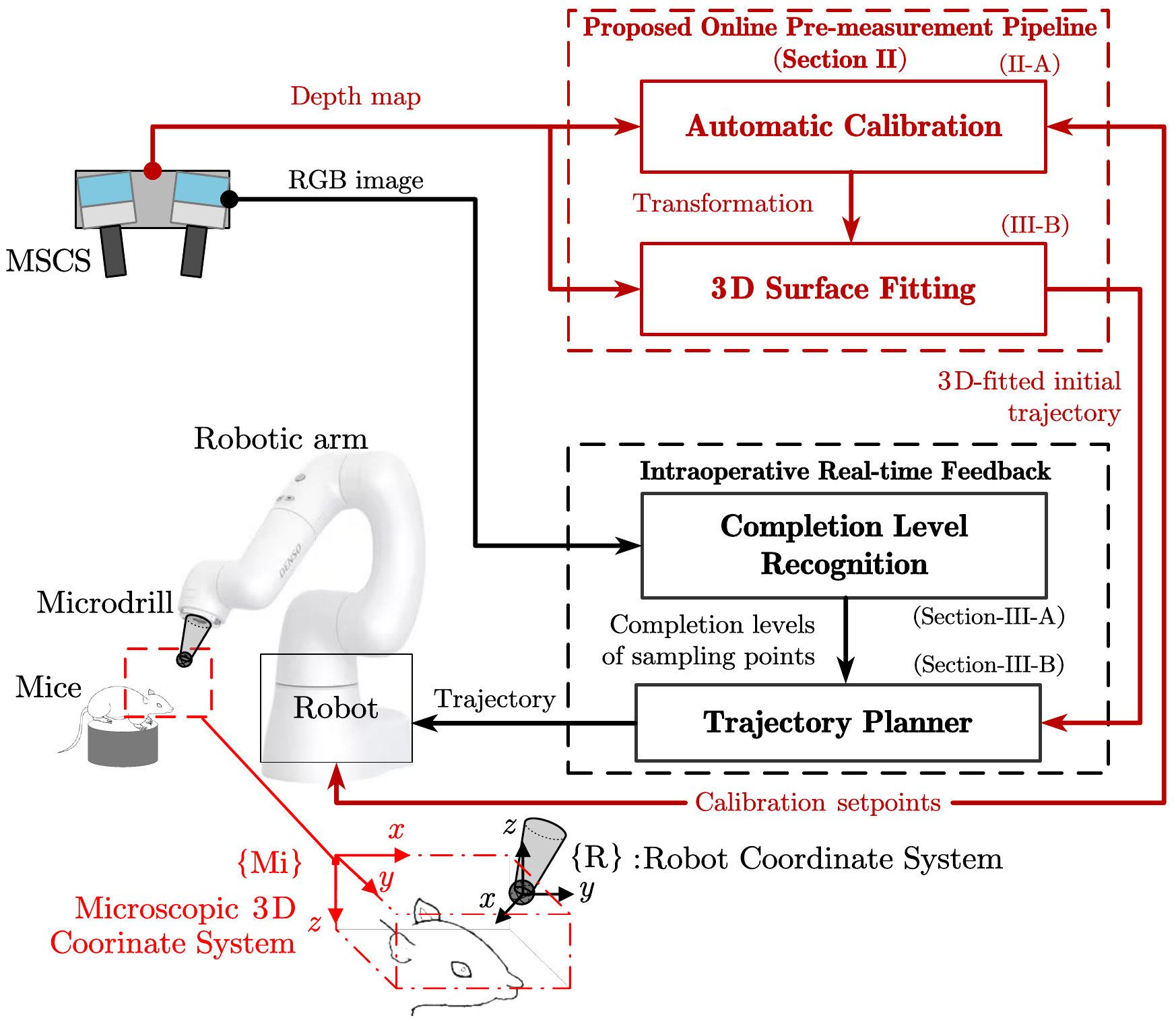}
\caption{End-to-end automation workflow of the improved autonomous robotic bone micro-milling system: the red part, Proposed Online Pre-measurement Pipeline, including Automatic Calibration (Section~\ref{sec:autocali}) and 3D Surface Fitting (Section~\ref{sec:initraje}), is the main contribution of this work and responsible for surface measurement before milling; the black part, Intraoperative Real-time Feedback, including Completion Level Recognition (Section~\ref{subsec:completionlevelrecognition}) and Trajectory Planner (Section~\ref{subsec:trajectoryplanning}), is improved from the original system \cite{zhao2023} and responsible for closed-loop control during milling. “Robot” refers to the robot motion controller, which controls the robot’s end effector (microdrill) to reach the specified coordinates. The robot coordinate system \{R\} and the microscopic 3D coordinate system \{Mi\} are illustrated at the bottom.}
\label{fig:overview}
\end{figure}

More importantly, in the aforementioned methods, manual assistance is required, either during the scanning process in offline measurement-based systems or during the initial trajectory adjustment in feedback-based systems. This reliance on the operator’s skill compromises the system’s autonomy, efficiency, and reproducibility, and therefore these systems cannot be regarded as truly autonomous.

In this study, we develop a robotic system capable of perceiving the uneven surface and heterogeneous thickness of the mouse skull and adapting in real time, enabling fully autonomous cranial window creation. To this end, we introduce a Microscopic Stereo Camera System (MSCS, see Fig.~\ref{fig:harware}) and propose an online pre-measurement pipeline. Unlike conventional offline measurement methods, which require a time-consuming scanning process and can only provide static anatomical information, this pipeline enables real-time reconstruction of the skull and automatically performs calibration and surface fitting, allowing rapid generation of an initial trajectory. By integrating this pipeline with the previous image-based intraoperative real-time feedback system, the improved automation workflow (see Fig.~\ref{fig:overview}) is expected to enhance the overall performance and eliminate the need for human assistance. Based on these advances, we present the world’s first autonomous robotic bone micro-milling system, which combines pre-measurement with real-time feedback, enabling rapid and precise end-to-end creation of mouse cranial windows. Experiments on euthanized mice demonstrate that, compared to the previous system, our approach substantially increases both speed and success rate, demonstrating its potential applicability in future in vivo studies. The comparison of this study's capabilities with existing works is presented in Table~\ref{table:1}. The main contributions of this work can be summarized as follows:

\begin{enumerate}
    \item An online pre-measurement pipeline was proposed, which, with the introduction of MSCS, starts from uncalibrated cameras to achieve automatic calibration and 3D surface fitting.
    \item An autonomous milling system was developed that synergizes the proposed pre-measurement pipeline with the image-based feedback in the prior research, achieving a rapid, precise, end-to-end automation workflow.
    \item Validation experiments on euthanized mice were conducted, demonstrating excellent performance and efficiency even compared with skilled human operators.
\end{enumerate}

\section{Proposed Online Pre-measurement Pipeline}\label{sec:method}

The proposed pipeline aims to rapidly and automatically perform online pre-measurement of the target surface and generate a 3D-fitted initial trajectory $\boldsymbol p_{ini}$ that adapts to the surface flatness for the initialization of the milling trajectory. $\boldsymbol p_{ini}$ is a $n$-dimensional vector, with each element defined by 
\begin{equation}
\boldsymbol p_{ini,i}\triangleq\begin{bmatrix}x_{ini,i} & y_{ini,i} & z_{ini,i}\end{bmatrix}^{\text{T}}\in\mathbb{R}^{3}, 
\label{piini}
\end{equation}

\noindent where $i=1, 2, ..., n$, and $n$ refers to the number of trajectory sampling points. The pipeline contains two parts, the Automatic Calibration (Section~\ref{sec:autocali}) that determines the transformation from the microscopic 3D coordinate system \{Mi\} to the robot coordinate system \{R\} (see Fig.~\ref{fig:overview}), and the 3D Surface Fitting (Section~\ref{sec:initraje}) that samples trajectory points on the target surface at appropriate locations and transforms them to \{R\}.

\subsection{Automatic Calibration} \label{sec:autocali}

\begin{figure}[!t]
\centering
\includegraphics[width=2.6in]{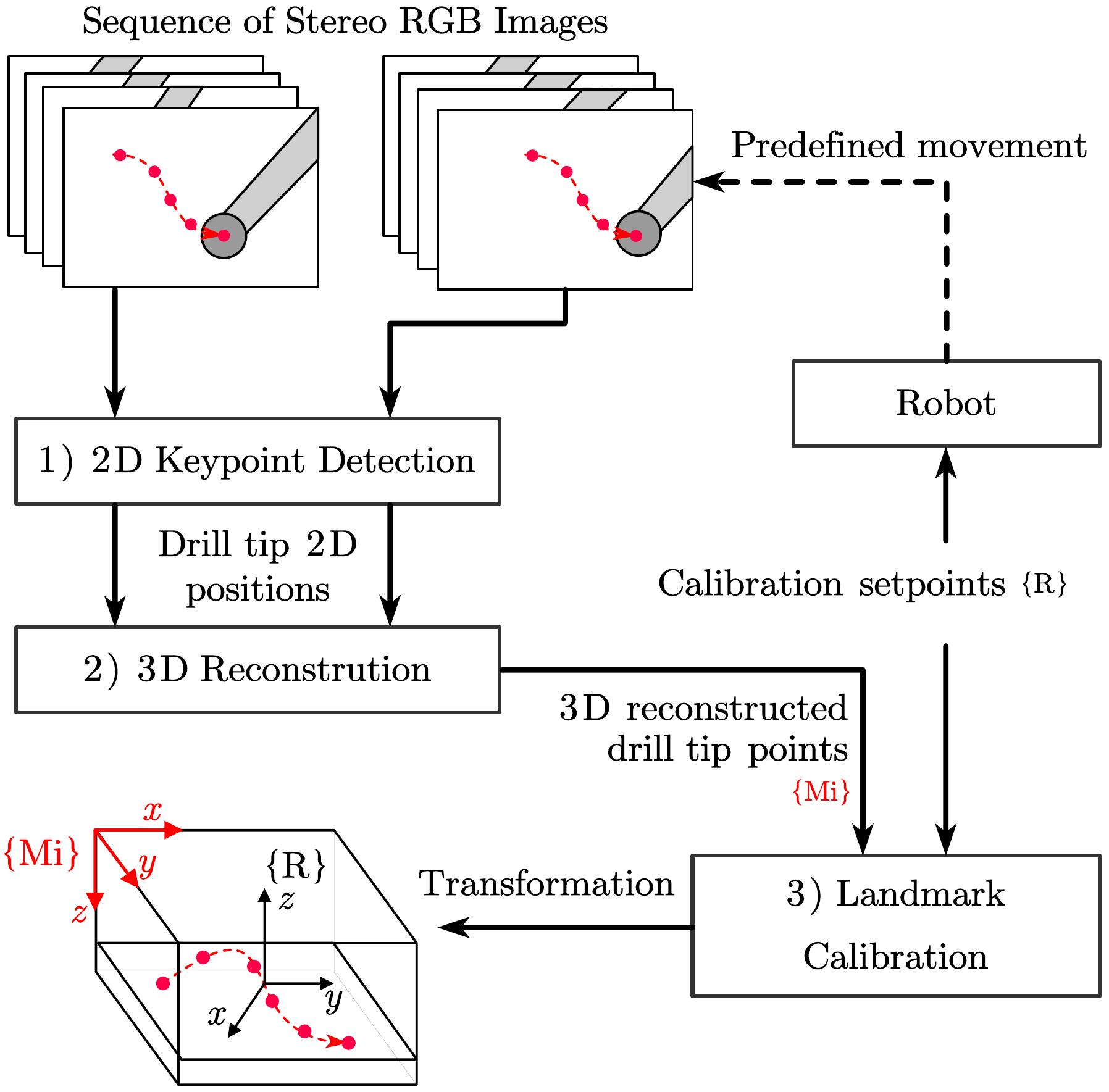}
\caption{The Automatic Calibration determines the rigid transformation matrix $\boldsymbol T_{\text{\{Mi\}}\rightarrow \text{\{R\}}}$ from the microscopic 3D coordinate system \{Mi\} to the robot coordinate system \{R\}. The robot executes a sequence of predefined movement commands to reach calibration setpoints, at each of which the 3D reconstructed drill tip point is obtained through 2D Keypoint Detection and 3D Reconstruction. Landmark Calibration then uses these 3D points and setpoints to calculate $\boldsymbol T_{\text{\{Mi\}}\rightarrow \text{\{R\}}}$.}
\label{fig:calibration}
\end{figure}

With the introduction of MSCS, the point clouds of the target surface can be reconstructed with respect to \{Mi\} using the model from the authors' previous work \cite{lin2024}. Since the milling trajectory must be planned relative to \{R\}, it is necessary to calibrate the transformation matrix from \{Mi\} to \{R\}. However, at the microscale, the limited workspace and lack of features on the microdrill make traditional hand-eye calibration methods infeasible, such as using Aruco markers \cite{garrido2014automatic}, checkerboard patterns \cite{zhang2000flexible}, or geometric landmarks \cite{manafifard2024review}. Therefore, in this section, we propose a keypoint detection-based Automatic Calibration, enabling rapid, real-time, and fully automated calibration, as shown in Fig.~\ref{fig:calibration}. The process includes 2D Keypoint Detection to detect 2D drill tip positions via a CNN,  3D Reconstruction to obtain 3D reconstructed drill tip points, and Landmark Calibration to solve the transformation between \{Mi\} and \{R\}.

\subsubsection{2D Keypoint Detection}\label{sec:2Dkeypointdetection}

To reconstruct the 3D points of the drill tip, its 2D projection coordinates in both the left and right images are required. This is primarily due to its small size, as directly accessing the reconstructed point clouds would lead to a suboptimal result. Therefore, a keypoint-detection CNN is utilized to obtain the drill tip 2D positions in both images. The CNN used is a minor adjustment of U-Net \cite{ronneberger2015u}, where the adjustments include changing the input ground truths from segmentation masks to confidence maps of drill tip localization. By this, the drill tip 2D positions $\boldsymbol p_l\triangleq\begin{bmatrix}x_l & y_l\end{bmatrix}^{\text{T}}\in\mathbb{R}^{2}$ in the left image and $\boldsymbol p_r\triangleq\begin{bmatrix}x_r & y_r\end{bmatrix}^{\text{T}}\in\mathbb{R}^{2}$ in the right image can be obtained by locating the local maximum from the predicted confidence map.

\subsubsection{3D Reconstruction}\label{sec:3Dreconstruction}

Based on the two detected drill tip 2D positions, the 3D position of the drill tip $\boldsymbol p_{d}^{\{\text{Mi}\}}\in\mathbb{R}^{3}$ with respect to \{Mi\} can be reconstructed by: 
\begin{equation}
\label{equ:lm}
\boldsymbol p_{d}^{\{\text{Mi}\}} =
\boldsymbol M_d 
\begin{bmatrix}
    \boldsymbol p_l \\ \boldsymbol p_r
\end{bmatrix}+\boldsymbol b_d,
\end{equation}

\noindent where $\boldsymbol M_d\in\mathbb{R}^{3 \times 4}$ is the coefficient matrix and $\boldsymbol b_d\in\mathbb{R}^{3 \times 1}$ is the bias, whose values can be calculated using the streamlined orthogonal projection model of disparity-depth mapping function following the methodology presented in \cite{lin2024}:
\begin{equation}
\label{Mdbd}
\boldsymbol M_d
=
\begin{bmatrix}
P_{\rho x} & 0 & 0& 0\\
P_{\rho y} & 0 & 0& 0\\
\frac{1}{h_{\rho}} & 0 & -\frac{1}{h_{\rho}} & 0
\end{bmatrix}, \ \
\boldsymbol b_d =
\begin{bmatrix}
-c_x P_{\rho x} \\
-c_y P_{\rho y}\\
d_e
\end{bmatrix},
\end{equation}

\noindent where $c_x,c_y$ (pixel) are the camera's principal point image coordinates, $P_{\rho x},P_{\rho y}$ (pixel/mm) respectively indicate the object's pixel sizes in the $x$ and $y$ directions, $d_e$ (mm) is the effective working distance of the stereo microscope setup, and $h_\rho $ (pixel/mm) denotes the depth response to unit disparity in reality. More details on the 3D reconstruction model and its calibration are provided in the Supplementary Materials. By substituting \eqref{Mdbd} into \eqref{equ:lm}, the 3D position of the reconstructed drill tip can be calculated.

\subsubsection{Landmark Calibration}\label{landmarkcalibration}

Landmark Calibration aims to determine the rigid transformation from \{Mi\} to \{R\} based on the corresponding landmarks. The transformation is defined as
\begin{equation}
\boldsymbol T_{\text{\{Mi\}}\rightarrow \text{\{R\}}}\triangleq
\begin{bmatrix}\boldsymbol R & \boldsymbol t \\ 0 & 1\end{bmatrix},
\label{Tmi2R}
\end{equation}

\noindent where $\boldsymbol{R}\in\mathbb{R}^{3 \times 3}$ is the rotation matrix and $\boldsymbol{t}\in\mathbb{R}^{3 \times 1}$ is the translation vector. 

To obtain the corresponding landmarks, $k$ calibration setpoints are sent to the robot to move the microdrill tip to these positions, whose coordinates with respect to \{R\} are represented as $\boldsymbol p_{i}^{\{\text{R}\}}\in\mathbb{R}^{3}, i=1,\cdots,k$. For each setpoint, two images are captured from left and right cameras in MSCS, and its 3D position with respect to \{Mi\} can be reconstructed by \eqref{equ:lm}. Thus, $k$ reconstructed drill tip points with respect to \{Mi\} can be obtained, represented as $\boldsymbol p_{i}^{\{\text{Mi}\}}\in\mathbb{R}^{3},i=1,\cdots,k$. Note that at least 3 non-collinear points ($k\geq3$) are required to uniquely determine the transformation. 

Finally, given the source landmarks $\boldsymbol p_{i}^{\{\text{Mi}\}}$ and the target landmarks $\boldsymbol p_{i}^{\{\text{R}\}}$, the transformation $\boldsymbol T_{\text{\{Mi\}}\rightarrow \text{\{R\}}}$ in \eqref{Tmi2R} can be calculated using the VTK's classic Landmark Transform Method\footnote{https://vtk.org/doc/nightly/html/classvtkLandmarkTransform.html}, thereby completing the Automatic Calibration.

\subsection{3D Surface Fitting}\label{sec:initraje}

\begin{figure}[!t]
\centering
\includegraphics[width=2.8in]{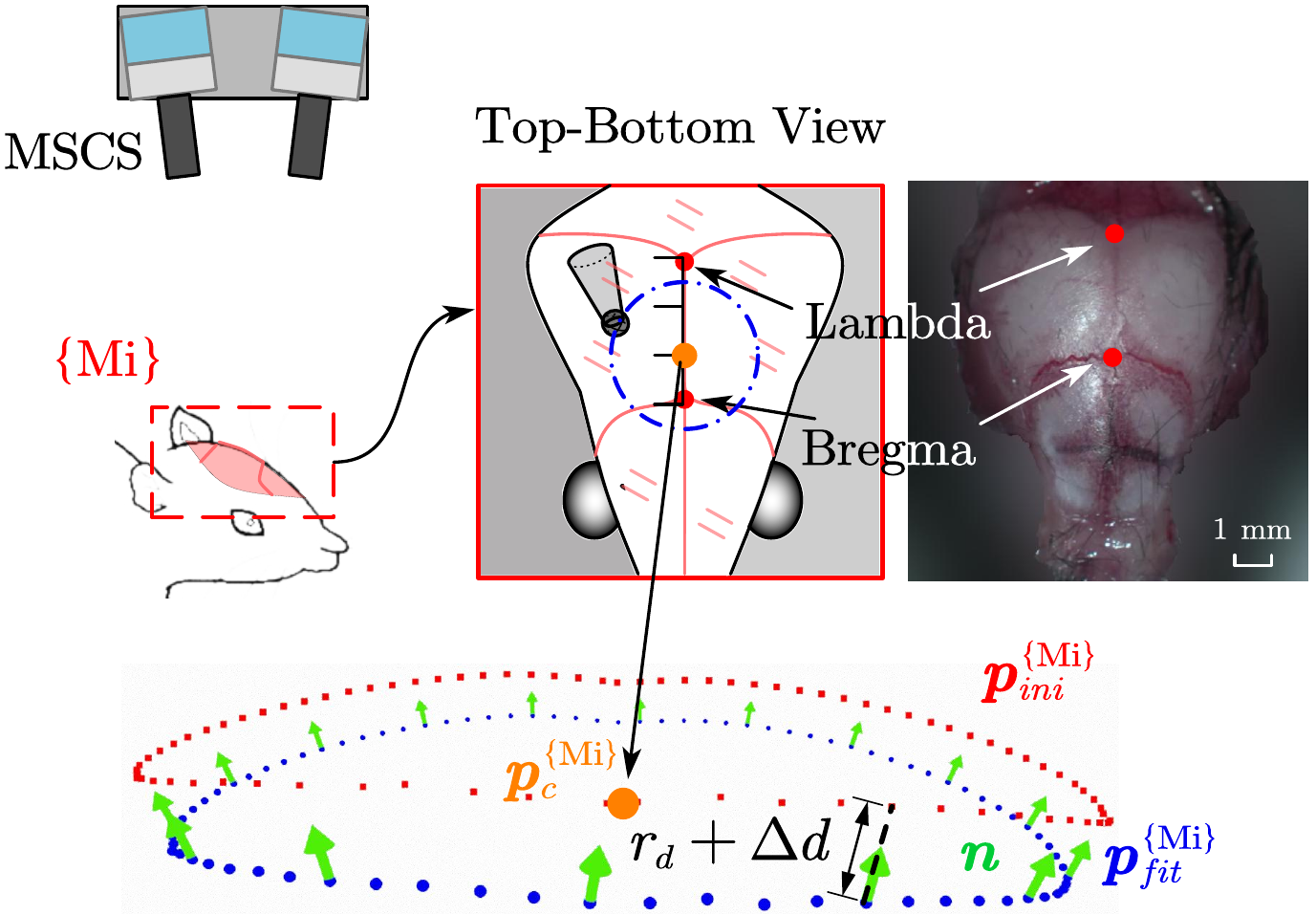}
\caption{
Schematic of 3D Surface Fitting. The bregma and lambda are detected through a CNN, and the milling center $\boldsymbol p_{c}^{\{\text{Mi}\}}$ is calculated by linearly weighting these two points. Subsequently, surface-fitted trajectory $\boldsymbol p_{fit}^{\{\text{Mi}\}}$ are sampled on the target surface based on $\boldsymbol p_{c}^{\{\text{Mi}\}}$ and radius $r$, and the 3D-fitted initial trajectory $\boldsymbol p_{ini}^{\{\text{Mi}\}}$ is obtained by offsetting $\boldsymbol  p_{fit}^{\{\text{Mi}\}}$ along the normal vectors $\boldsymbol n$ by a distance $r_d+\Delta d$.}

\label{fig:init_trajectory}
\end{figure}

In this section, the goal is to generate an initial trajectory that is 3D-fitted to the target surface. The center point of the trajectory is determined through the Trajectory Center Localization, while the trajectory points are distributed along the circular path and further offset through the 3D-fitted Initial Trajectory Generation. A schematic is illustrated in Fig.~\ref{fig:init_trajectory}.

\subsubsection{Trajectory Center Localization} \label{sec:trajectorycenterloc}

In the actual experiment of mouse cranial window creation, the center of the circular milling trajectory is always located between two specific anatomical points on the skull: bregma and lambda. Inspired by the method in \cite{zhou2020automatically}, which uses a CNN to automatically detect bregma and lambda, we applied a U-Net-based CNN with the same architecture as in Section~\ref{sec:2Dkeypointdetection} for the detection of these two points. 

In practice, the ratio $l$ of the distance between the trajectory center and bregma to the distance between bregma and lambda is determined based on the radius of the milling trajectory $r$ and the size of the mouse. Therefore, the 2D position of the trajectory center in the input image can be obtained by proportionally weighting the detected positions of bregma and lambda. Then, due to the high-quality reconstruction of the skull surface, the 3D localization of the trajectory center $\boldsymbol p_{c}^{\{\text{Mi}\}}\triangleq\begin{bmatrix}x_c & y_c & z_c\end{bmatrix}^{\text{T}}\in\mathbb{R}^{3}$ with respect to \{Mi\} can be performed by accessing its 2D position in the RGB image from either camera and the reconstructed point cloud of the target surface (detailed in \cite{lin2024}).

\subsubsection{3D-fitted Initial Trajectory Generation}

Considering that the expected trajectory is circular when projected onto the $x\text{--}y$ plane, by combining the obtained trajectory center $\boldsymbol p_{c}^{\{\text{Mi}\}}$ and the milling trajectory radius $r$, a surface-fitted trajectory $\boldsymbol p_{fit}^{\{\text{Mi}\}}$ with respect to \{Mi\} can be sampled on the reconstructed point cloud. The $x$ and $y$ coordinates of its $i$-th sampling point $\boldsymbol p_{fit,i}^{\{\text{Mi}\}}\triangleq\begin{bmatrix}x_{fit,i} & y_{fit,i} & z_{fit,i}\end{bmatrix}^{\text{T}}\in\mathbb{R}^{3}$ can be calculated by:

\begin{equation}
\begin{aligned}
x_{fit,i} &= r\cos{\theta_i} + x_c \\
y_{fit,i} &= r\sin{\theta_i} + y_c,
\end{aligned}
\label{pfit}
\end{equation}

\noindent where $\theta_{i}=2{\pi}i/n, i=1,\cdots,n$. Then $z_{fit,i}$ can be obtained by combining $x_{fit,i}, \ y_{fit,i}$ and the reconstructed point cloud of the target surface.

In order to avoid contact with the surface before milling starts, an offset should be set for all sampling points along their normal directions, where the normal vector of each point $\boldsymbol n_{i}\in\mathbb{R}^{3}, i=1,\cdots,n$ can be calculated from the reconstructed point cloud. As a result, the 3D-fitted initial trajectory $\boldsymbol p_{ini}^{\{\text{Mi}\}}$ with respect to \{Mi\} can be obtained, where each element can be calculated by
\begin{equation}
\boldsymbol p_{ini,i}^{\{\text{Mi}\}} = \boldsymbol p_{fit,i}^{\{\text{Mi}\}} + (r_d+\Delta d)\cdot\boldsymbol n_i.
\label{pini}
\end{equation}

\noindent where $r_d$ is the radius of the drill tip, and $\Delta d$ is a manually set offset for safety concerns, taking into account both preliminary measurements of the 3D Surface Fitting module and a conservative margin to accommodate measurement uncertainty and possible unmodeled variations.

Lastly, utilizing the transformation matrix output from Automatic Calibration by \eqref{Tmi2R}, we can acquire the $n$-dimensional vector of transformed 3D-fitted initial trajectory $\boldsymbol p_{ini}$ with respect to \{R\}, where each element is defined by \eqref{piini} and can be calculated by 
\begin{equation}
\begin{bmatrix}\boldsymbol p_{ini,i} \\ 1\end{bmatrix}
 = \boldsymbol T_{\text{\{Mi\}}\rightarrow \text{\{R\}}} {\cdot} 
\begin{bmatrix}\boldsymbol p_{ini,i}^{\{\text{Mi}\}} \\ 1\end{bmatrix}.
\label{pcr}
\end{equation}

The result of \eqref{pcr} is output to the Trajectory Planner (Section~\ref{subsec:trajectoryplanning}) for the initialization of the milling trajectory. 

\section{End-to-end Automation Workflow}\label{sec:material}

With the integration of the proposed online pre-measurement pipeline (Section~\ref{sec:method}), we improve the autonomous robotic bone micro-milling system and achieve an end-to-end automation workflow (see Fig.~\ref{fig:overview}). In this section, we briefly introduce the intraoperative real-time feedback loop in the workflow, consisting of Completion Level Recognition (Section~\ref{subsec:completionlevelrecognition}) and Trajectory Planner (Section~\ref{subsec:trajectoryplanning}), which are modified from the authors’ previous system \cite{zhao2023} and ensure the system's adaptive control after the online pre-measurement phase is completed and milling begins.

\subsection{Completion Level Recognition}\label{subsec:completionlevelrecognition}

The Completion Level Recognition aims to monitor the milling completion degree along the trajectory in real-time using a CNN based on the input RGB images. The CNN follows the same DSSD-inspired \cite{fu2017dssd} architecture as the previous system \cite{zhao2023}, but is retrained on a new mouse skull dataset. The CNN outputs a bounding box for milling area detection and a heatmap image for pixel-wise completion level prediction. The network was trained on an augmented dataset derived from 587 manually collected and annotated images (a total of 18,784 training images) from videos of teleoperated robotic cranial window creation experiments conducted on 6 euthanized mice, achieving 77.6 mAP for the detection task and 24.32\% MAPE for the completion prediction task, with a processing speed of 72 Hz. More training details (network architecture, dataset creation, data annotation, loss function, training condition, extended evaluation, and output example) are available in the Supplementary Materials.

It is important to note that although the 24.32 \% MAPE may not seem ideal, this is due to the strictness of pixel-level completion prediction. In actual milling experiments, the system is relatively insensitive to variations at low completion levels and only needs to accurately capture the moment of penetration. The color and light response changes of the skull at penetration coincidentally enable the CNN to predict this critical moment with reasonable accuracy. Nevertheless, prediction errors can still occur, potentially causing experimental failures, which will be discussed in detail in Section~\ref{sec:resultanddescussion}.

Through post-processing and sampling of the output image, an $n$-dimensional vector $\boldsymbol{c}$ can be obtained, where each element denotes the penetration percentage at the $i$-th sampling point along the real-time-updated trajectory, defined by

\begin{equation}
\mathbb{R}\ni c_{i}\triangleq c_{i}\left(t\right)\in\left[0,1\right],
\label{ci}
\end{equation}

\noindent where $i=1, 2, ..., n$. Note that $c_{i}=0$ means untouched and $c_{i}=1$ means completely penetrated. $\boldsymbol{c}$ is output to the Trajectory Planner for real-time trajectory updates.

\subsection{Trajectory Planner} \label{subsec:trajectoryplanning}

\begin{figure}[!t]
\centering
\includegraphics[width=3.2in]{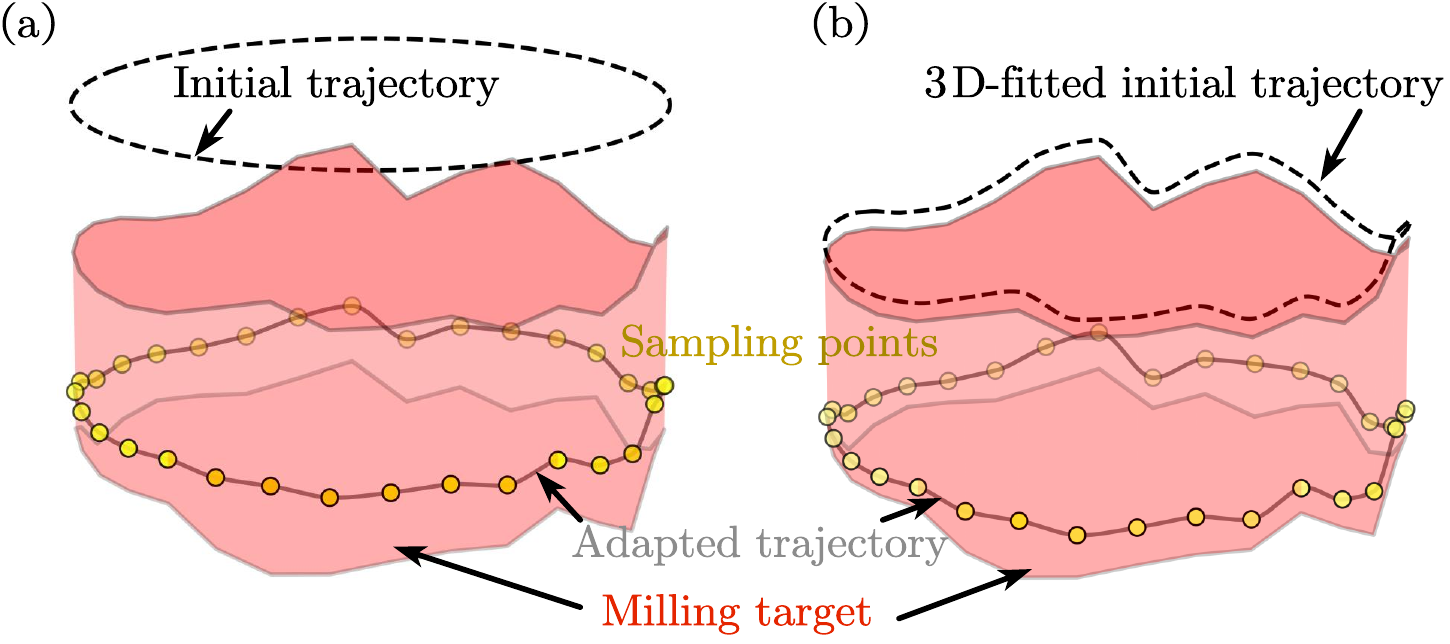}
\caption{Schemas of (a) the Trajectory Planner in the previous system, and (b) the Trajectory Planner in this study.}
\label{fig:material}
\end{figure}

The Trajectory Planner is responsible for initializing, updating and smoothing the milling trajectory. The updating process remains consistent with the previous system, employing a multimodal velocity damper to control the descent speed of the drill bit at the $i$-th sampling point to be proportionally reduced according to its completion level $c_{i}$, which is real-time output from the Completion Level Recognition and defined by \eqref{ci}.

For the initialization process, in the previous system, due to the absence of direct surface measurements, the $z$-coordinates of all sampling points are initialized to 0 (see Fig.~\ref{fig:material}-(a)). In this study, with the integration of MSCS, a 3D-fitted initial trajectory $\boldsymbol p_{ini}$ is available, output from the 3D Surface Fitting (Section~\ref{sec:initraje}) and defined by \eqref{piini}. The trajectory features a non-zero $z$-axis coordinate at each sampling point to pre-fit the uneven surface (see Fig.~\ref{fig:material}-(b)), enabling the microdrill to contact the target surface rapidly and spatially consistently.

The smoothing process remains the same as in the previous system, where a constrained cubic spline interpolation method \cite{kruger2003constrained} is used to generate a continuous trajectory based on the $n$ discrete sampling points, thereby matching the high-precision encoders used in the robotic system \cite{marinho2024}. Additionally, we hypothesize that if all $n$ sampling points are completely penetrated, the entire smoothed trajectory will be sufficiently milled. Therefore, when $c_i = 1 \ \ \forall i = 1, \cdots, n$, a stop signal will be triggered and sent to the controller to automatically halt the robot's movement and the drill's rotation, marking the completion of the milling process. More details on the mathematical description of the Trajectory Planner are available in the Supplementary Materials.

\section{Experiments}

In this section, experiments are first conducted to validate the effectiveness of the Automatic Calibration and the 3D Surface Fitting, which together constitute the proposed online pre-measurement pipeline. Subsequently, further experiments are conducted to evaluate the performance of the entire autonomous robotic bone micro-milling system. More details on the hardware setup and software implementation applied in the experiments are available in the Supplementary Materials.

\subsection{Parameter selection}

To implement the proposed method for validations and experiments, the parameters mentioned in Section~\ref{sec:method} and Section~\ref{sec:material} are selected based on camera calibration, actual measurements, safety considerations, and ease of calculation: the number of trajectory sampling points $n=32$, the intrinsic parameters of cameras $c_x=480 \ \mathrm{pixel}$, $c_y=270 \ \mathrm{pixel} $, $P_{\rho x}=0.03947 \ \mathrm{pixel/mm}$, $P_{\rho y}=0.03894 \ \mathrm{pixel/mm}$, $d_e=0$, and $h_\rho=6.4 \ \mathrm{pixel/mm}$, the ratio of the distance between the trajectory center and bregma to the distance between bregma and lambda $l=1/3$, the radius of the milling trajectory $r= 2 \ \mathrm{mm}$, the radius of the drill tip $r_d=0.5 \ \mathrm{mm}$, the manually set offset $\Delta d=0.5 \ \mathrm{mm}$, the number of calibration setpoints $k=5$ and their coordinates $\boldsymbol p_{1}^{\{\text{R}\}}= \begin{bmatrix}0 & -20 & 0\end{bmatrix}^{\text{T}}, \ \boldsymbol p_{2}^{\{\text{R}\}}=\begin{bmatrix}10 & -20 & 0\end{bmatrix}^{\text{T}}, \ \boldsymbol p_{3}^{\{\text{R}\}}=\begin{bmatrix}5 & -10 & 0\end{bmatrix}^{\text{T}}, \ \boldsymbol p_{4}^{\{\text{R}\}}=\begin{bmatrix}10 & 0 & 0\end{bmatrix}^{\text{T}}, \ \boldsymbol p_{5}^{\{\text{R}\}}=\begin{bmatrix}0 & 0 & 0\end{bmatrix}^{\text{T}} \ \mathrm{mm}$. Additionally, the resolution of the images captured by the MSCS is set to $540 \times960$ to meet the requirement for each module.

\subsection{Experiment 1: Validation of Automatic Calibration} \label{joint_evaluation}

\begin{figure}[!t]
\centering
\includegraphics[width=2.7in]{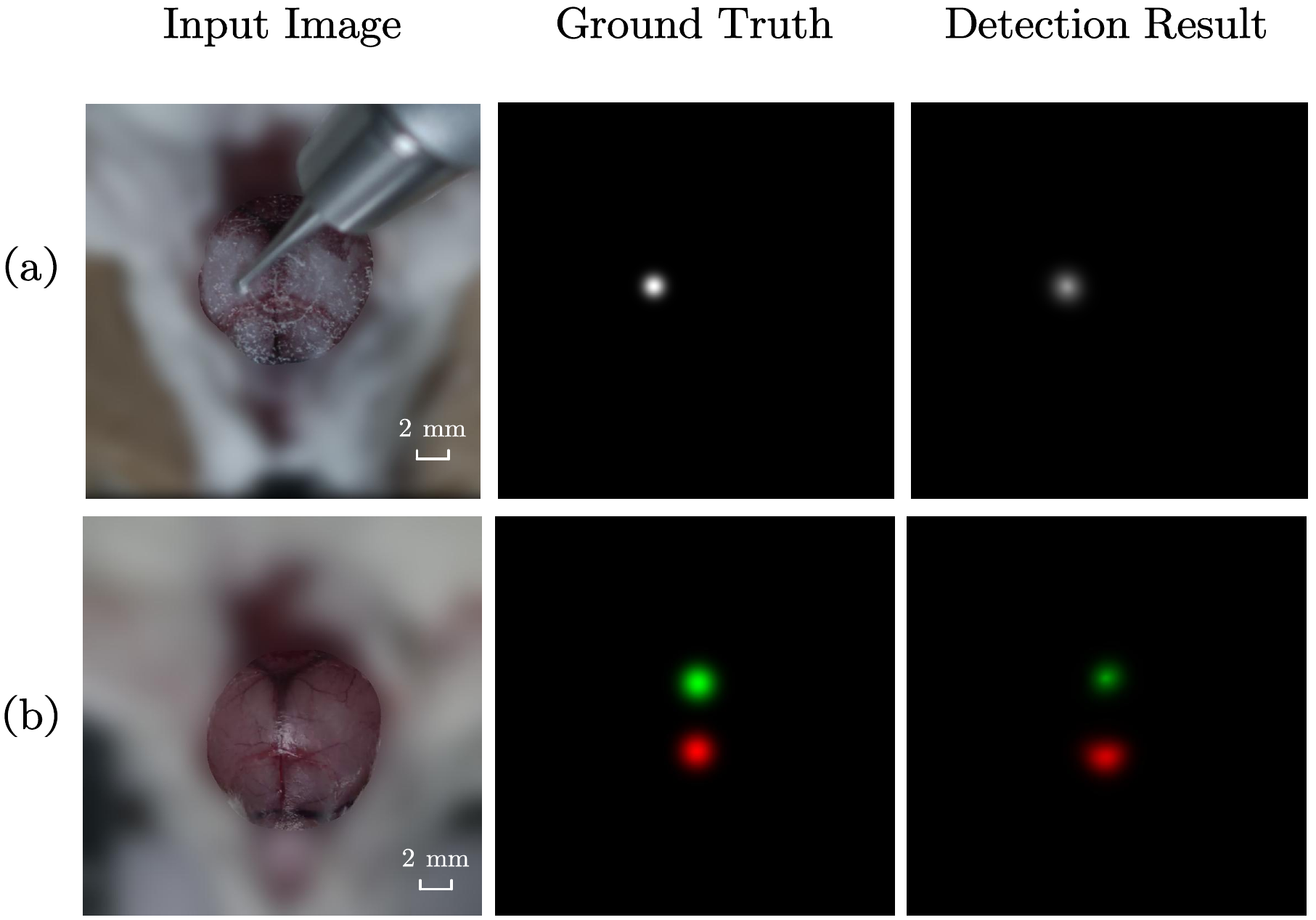}
\caption{Examples of the input image, ground truth, and detection result for the keypoint detection CNNs on (a) the drill tip and (b) bregma and lambda. The ground truth and detection result images are confidence maps, where pixels with brighter colors indicate a higher probability of being keypoint locations (while: drill tip, red: bregma, green: lambda).}
\label{fig:example}
\end{figure}

As described in Section~\ref{sec:autocali}, the accuracy of Automatic Calibration is affected by the accuracy of 2D Keypoint Detection, 3D Reconstruction, and Landmark Calibration. Considering that the 3D Reconstruction depends on the MSCS, whose accuracy has been evaluated in prior research by the present authors \cite{lin2024}, in Experiment 1, we focus on evaluating 2D Keypoint Detection and Landmark Calibration.

\subsubsection{Training of 2D Keypoint Detection model}

A total of 3,294 images, each showing the drill tip within the field of view, were captured from videos of teleoperated cranial window experiments on 10 euthanized mice. No additional ethical approval was needed as the mice were reused from another study. For each image, a confidence map was generated by applying a 2D Gaussian to the manually annotated drill tip position, serving as ground truth (an example is shown in Fig.~\ref{fig:example}-(a)). Data augmentation, including blur, noise, color adjustment, geometric transformation, and resolution changes, expanded the dataset to 16,394 images, with 14,754 for training, 820 for validation, and 820 for testing.

The models were trained for 30 epochs using the Adam optimizer with Mean Square Error (MSE) loss. The learning rate was initially set to $1\times10^{-3}$, then reduced to $1\times10^{-4}$ for epochs 11 to 20, and finally to $1\times10^{-5}$ for epochs 21 to 30, with a batch size of 4. Training was implemented in PyTorch 1.8.0 with CUDA 11.1 on Python 3.8, running on Ubuntu 20.04 with an NVIDIA Quadro P6000 GPU.

\subsubsection{Joint evaluation metrics}
\label{subsubsec:jointevaluationmetrics}

Firstly, we randomly generate $m=20$ validation setpoints $\boldsymbol p_{val,i}^{\{\text{R}\}}\in\mathbb{R}^{3},i=1,\cdots,m$ within the reconstruction space of MSCS ($x\in[0,10], \ y\in[-20,0], \ z\in[0,10] \ \mathrm{mm}$) and send them to the robot so that the drill tip reaches these points. MSCS captures $m$ images each from left and right camera, outputting $m$ drill tip 2D positions each $\boldsymbol p_{l,i}, \ \boldsymbol p_{r,i}\in\mathbb{R}^{2},i=1,\cdots,m$ via 2D Keypoint Detection (see Section~\ref{sec:2Dkeypointdetection}). Then we manually annotate the left and right images, obtaining $m$ ground truths each $\boldsymbol p_{l,i}^{(gt)}, \ \boldsymbol p_{r,i}^{(gt)}\in\mathbb{R}^{2},i=1,\cdots,m$. Through 3D Reconstruction (see Section~\ref{sec:3Dreconstruction}), $m$ points of the 3D reconstructed drill tip with respect to \{Mi\} can be obtained. Then, by transforming these points back to \{R\} using the transformation calculated from Landmark Calibration (see Section~\ref{landmarkcalibration}), we can obtain $\boldsymbol p_{trans,i}^{\{\text{R}\}}\in\mathbb{R}^{3},i=1,\cdots,m$. The accuracies of 2D Keypoint Detection and Landmark Calibration are respectively evaluated by the Root Mean Square Error (RMSE) between $\boldsymbol p_{l,i}, \ \boldsymbol p_{r,i}$ and $\boldsymbol p_{l,i}^{(gt)}, \ \boldsymbol p_{r,i}^{(gt)}$, and between $\boldsymbol p_{trans,i}^{\{\text{R}\}}$ and $\boldsymbol p_{val,i}^{\{\text{R}\}}$. 

Automatic Calibration mainly provides a basis for rapidly generating the initial surface-fitting trajectory, so its accuracy requirement prioritizes efficiency. Therefore, based on preliminary experiments and practical experience, the error in 2D keypoint detection is required not to exceed 10 pixels, and the error of Landmark Calibration is required not to exceed 1 mm.

\subsubsection{Result and discussion}

\begin{figure}[!t]
\centering
\includegraphics[width=2.7in]{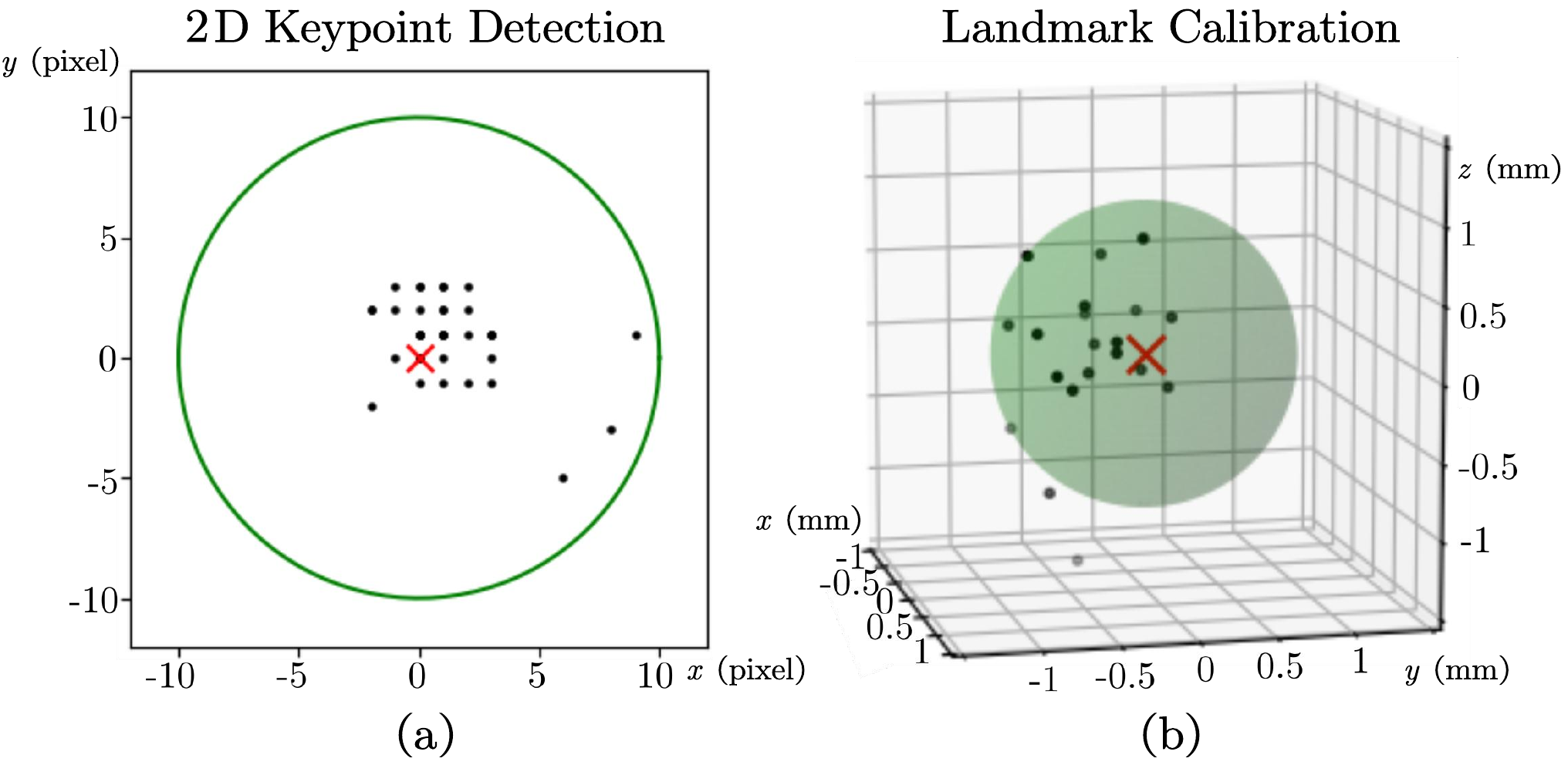}
\caption{Evaluation results of (a) 2D Keypoint Detection and (b) Landmark Calibration on 20 validation setpoints. The black dots, red cross and green circle or sphere respectively represent (a) the centralized $\boldsymbol p_{l,i}, \  \boldsymbol p_{r,i}$, the centralized $\boldsymbol p_{l,i}^{(gt)}, \ \boldsymbol p_{r,i}^{(gt)}$, the accuracy requirement of 10 pixels; (b) the centralized $\boldsymbol p_{trans,i}^{\{\text{R}\}}$, the centralized $\boldsymbol p_{val,i}^{\{\text{R}\}}$, and the accuracy requirement of 1 mm.}
\label{fig:ex1result}
\end{figure}

The evaluation results, shown in Fig.~\ref{fig:ex1result}, indicate an RMSE of 3.18 pixels for 2D Keypoint Detection and 0.77 mm for Landmark Calibration, both meeting the required accuracy. The validation setpoints and their reconstructed results are provided in the Supplementary Materials.

Specifically, as shown in Fig.~\ref{fig:ex1result}-(a), all 20 validation points exhibited 2D detection errors within the threshold, demonstrating good accuracy and robustness of the model. For Landmark Calibration, although the 0.77 mm RMSE satisfies the requirement, it exceeds the safety offset $\Delta d$ (0.5 mm), which theoretically increases the risk of the drill contacting the skull surface before milling. Moreover, as shown in Fig.~\ref{fig:ex1result}-(b), 4 points exceeded the error threshold. Further analysis revealed that these points were all located near the image periphery. This is likely due to the relatively large tilt angle ($\phi$), i.e., deviation from the vertical direction, which reduces the stability of 3D reconstruction in peripheral regions. In practice, trajectory sampling points are concentrated near the image center to ensure reconstruction accuracy. Among the 20 randomly generated validation points, 7 were located at the image periphery. After manually excluding these points, the RMSE of the remaining points decreased to 0.31 mm, which is below $\Delta d$, indicating that the actual calibration accuracy in the central region is significantly higher than the overall random-test performance. Future work will focus on improving 3D reconstruction accuracy by employing higher-resolution cameras and more robust models, thereby enhancing spatial localization performance.

\subsection{Experiment 2: Validation of 3D Surface Fitting}

As described in Section~\ref{sec:initraje}, the accuracy of 3D Surface Fitting is affected by the accuracy of Trajectory Center Localization and 3D-fitted Initial Trajectory Generation. As the performance of 3D-fitted Initial Trajectory Generation is based on the reconstruction of the target surface, which has been evaluated in \cite{lin2024}, and the transformation obtained via Landmark Calibration, which has been evaluated in Section~\ref{joint_evaluation}, in Experiment 2, we focus on the evaluation of Trajectory Center Localization.

\subsubsection{Training of CNN model}

A dataset consisting of 523 images that satisfy the requirement that the mouse skulls in the images remain unmilled was captured from videos of the same 10 euthanized mice mentioned in Section~\ref{joint_evaluation}, as well as 82 images from a publicly available dataset created by \cite{zhou2020automatically}. The creation of ground truths and data augmentation are the same as Section~\ref{joint_evaluation}, resulting in a total of 8,774 images, with 7,019 for training, 877 for validation, and 878 for testing (an example is shown in Fig.~\ref{fig:example}-(b)). Training conditions and parameters are consistent to Section~\ref{joint_evaluation}.

\subsubsection{Metrics}

Considering that the trajectory center is calculated by linearly weighting the detected bregma and lambda, with the detection errors of these two points considered independent, the error of Trajectory Center Localization can also be calculated by linearly weighting the detection RMSE of bregma and lambda. Same as Section~\ref{joint_evaluation}, we require the error not to exceed 10 pixels.

\subsubsection{Result and discussion}

In the test dataset, we observed an RMSE of 1.19 pixels for bregma detection and 1.73 pixels for lambda detection. The RMSE for lambda was higher than that for bregma, possibly because bregma is more distinguishable than lambda, even for humans. By applying linear weighting to integrate the information from both, the RMSE of the trajectory center was 1.37 pixels, meeting the accuracy requirement.

\subsection{Experiment 3: Validation of the autonomous robotic bone micro-milling system}

\begin{figure}[!t]
\centering
\includegraphics[width=2.8in]{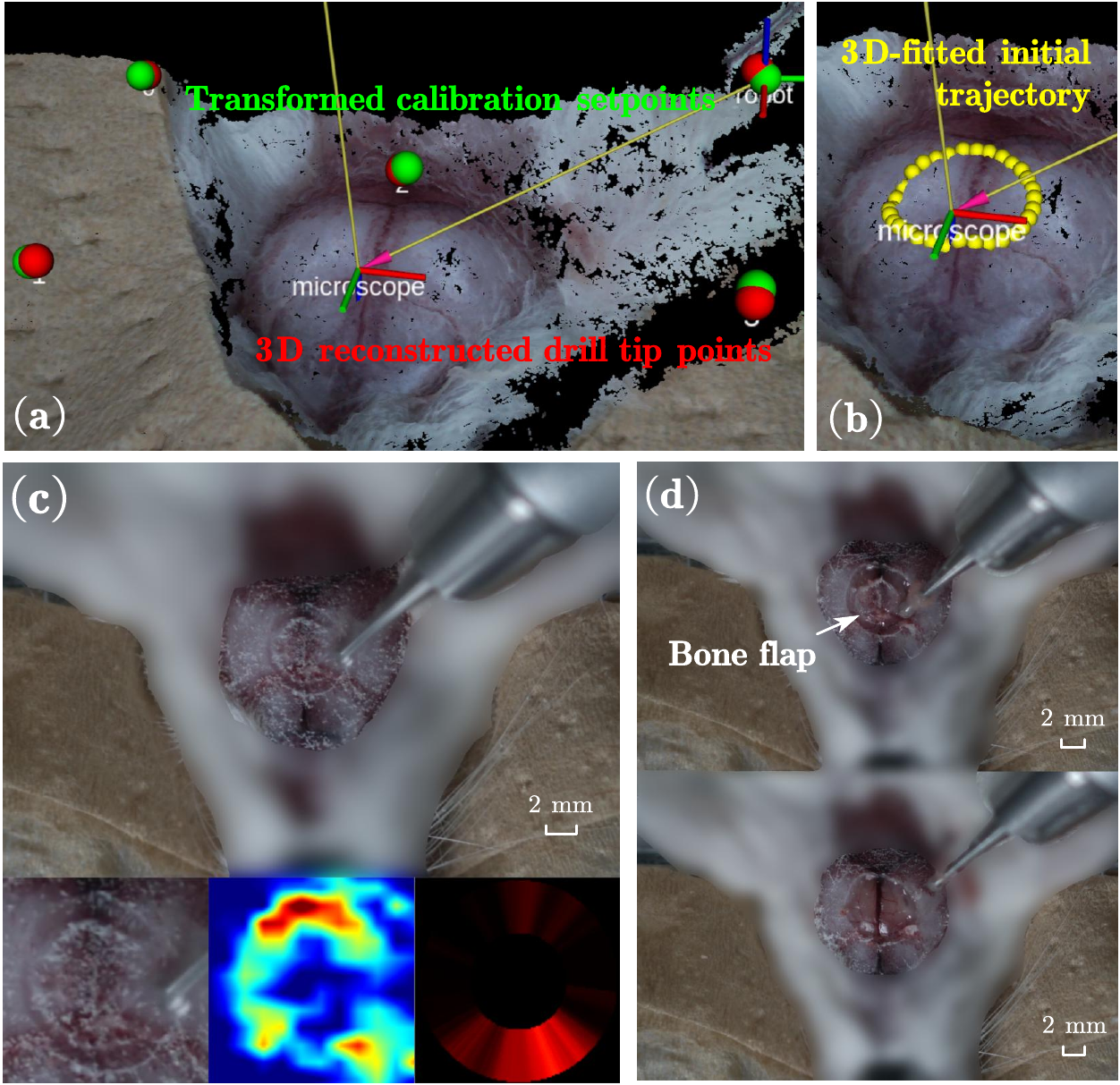}
\caption{The automation flow of an experimental case.
(a) The result of Automatic Calibration (Section~\ref{sec:autocali}). (b) The result of 3D Surface Fitting (Section~\ref{sec:initraje}). (c) The input and output of the Completion Level Recognition (Section~\ref{subsec:completionlevelrecognition}). (d) The result after the milling process was completed and after the bone flap was manually removed.}
\label{fig:ex3result}
\end{figure}

Following the end-to-end automation workflow shown in Fig.~\ref{fig:overview}, with its hardware setup shown in Fig.~\ref{fig:harware}, autonomous cranial window creation experiments were conducted on euthanized mice to validate the feasibility of the proposed method. 

\subsubsection{Animal model}

All experiments were performed on mice euthanized with carbon dioxide, approved by the Animal Care and Use Committee, Graduate School of Medicine, The University of Tokyo (Approval No.: A2023M042-04). The experiments were conducted in compliance with the Fundamental Guidelines for Proper Conduct of Animal Experiment and Related Activities in Academic Research Institutions (2006), MEXT Japan, and Guidelines for Proper Conduct of Animal Experiments (2006), Science Council of Japan. 7 female NOD.CB17-Prkdc\textsuperscript{scid}/J mice, aged 6 to 8 weeks and weighing 20-25 g, are used in the experiments. 

\subsubsection{Metrics}

Milling success and time are used to evaluate the performance. Success is defined as the bone flap (see Fig.~\ref{fig:ex3result}-(d)) within the milling trajectory being removable manually with tweezers after the milling is completed, without damaging the membrane beneath the skull \cite{zhao2023}.

\subsubsection{Result and discussion}\label{sec:resultanddescussion}

7 experiments were conducted, and 6 were successful, resulting in a success rate of 85.7\,\%. The average required time for successful milling was $2.1\pm 1.8$ minutes. Compared with the 80\,\% of success rate and the $16.8\pm 2.5$ minutes of milling time using the previous system \cite{zhao2023} evaluated on the eggshell model, the feasibility and efficiency of the improved autonomous robotic bone micro-milling scheme were initially proved on euthanized mice. The overall improvement in milling performance highlights the importance of the proposed online pre-measurement pipeline, which effectively eliminates idle time caused by the lack of pre-measured anatomical information of the skull. Crucially, this substantial performance gain was achieved in the more challenging context of the non-uniform, biological mouse skull, in contrast to the simpler, uniform eggshell model used in the prior study. Fig.~\ref{fig:ex3result} shows a successful milling case. As expected, the MSCS improved surface perception, increasing the success rate and reducing milling time compared to the previous system that relied solely on real-time feedback.

Furthermore, compared with the results of teleoperated robotic cranial window creation mentioned in Section~\ref{joint_evaluation} ($13.7\pm 6.6$ minutes, 3 skilled and 4 novice human operators, 10 euthanized mice) and manual cranial window creation \cite{navabi25} ($8.0\pm 2.1$ minutes, 2 skilled human operators, 6 anesthetized live mice), the system is demonstrated with excellent efficiency, repeatability, and consistency.

In the 7 experiments, the only failure case was primarily caused by an error in the Completion Level Recognition. In this case, the completion level of the area that had already been completely penetrated, which should have been 1, was incorrectly recognized as being below 1, causing the drill tip to continue descending. Consequently, the underlying membrane was damaged and the automatic milling process was stopped manually. We consider that the accuracy of the current recognition CNN, relying solely on RGB images, limits the upper bound of the system’s success rate. Future work will focus on improving the recognition accuracy of the CNN or incorporating additional signals to assist image processing.

It is noteworthy that a potential failure mode in previous systems did not occur in our experiments, where prolonged milling duration caused the drill to repeatedly contact already penetrated membrane areas, eventually leading to rupture. By shortening the overall milling duration, we effectively reduced the drill’s dwell time on these vulnerable areas, thereby lowering the risk of membrane damage.

\section{Discussion and Conclusions}

In this paper, we propose an online pre-measurement pipeline of automatic calibration and 3D surface fitting, and integrate it into an autonomous robotic bone micro-milling system, enabling an end-to-end autonomous cranial window creation workflow capable of rapidly, in real-time, and accurately perceiving and adapting to the uneven surface and non-uniform thickness of bone tissue. Crucially, this work goes beyond the framework of the previous system by eliminating the need for manual assistance and addressing the main limitations, resulting in a system that is not only truly autonomous but also achieves comprehensive improvements in performance and efficiency. In the experiments, we validated the system's excellent performance and efficiency by successfully performing autonomous cranial window creation on euthanized mice, achieving faster, more accurate and more stable results compared to skilled human operators.

Future work includes further improving the system performance and applying it to anesthetized live mice.  Beyond the mouse cranial window creation, the proposed framework holds strong potential for generalization to other microsurgical bone milling tasks requiring high precision and tissue preservation. The integration of pre-measurement and real-time feedback enables adaptive control over heterogeneous biological structures, suggesting a general robotic strategy for microscale bone processing and possible extensions to broader medical applications such as endoscopic or neurosurgical procedures.

\section*{Acknowledgment}

We express our special gratitude to Dr. Saúl Alexis Heredia Pérez, Dr. Murilo Marques Marinho, and the surgeons at the Institute of Science Tokyo for their advice and help.

\addtolength{\textheight}{-12cm}   





\bibliographystyle{IEEEtran}
\bibliography{references}

\end{document}